\documentclass{article}
\usepackage{spconf,amsmath,graphicx}
\usepackage{xcolor}

\usepackage{hyperref}
\usepackage{url}
\usepackage{stfloats} 
\usepackage[utf8]{inputenc} 
\usepackage[T1]{fontenc}    
\usepackage{hyperref}       
\usepackage{url}            
\usepackage{booktabs}       
\usepackage{amsfonts}       
\usepackage{nicefrac}       
\usepackage{microtype}      
\usepackage{xcolor}         
\usepackage{amsthm} 
\usepackage{bm} 
\usepackage{algpseudocode, algorithm}
\usepackage{etoolbox}
\patchcmd{\thmhead}{(#3)}{#3}{}{}

\newtheorem{theorem}{Theorem}[section]

\newtheorem{definition}{Definition}[section]

\title{Toward Asymptotic Optimality: Sequential Unsupervised\\Regression of Density Ratio for Early Classification}
\name{Akinori F. Ebihara, Taiki Miyagawa, Kazuyuki Sakurai, Hitoshi Imaoka}
\address{NEC Corporation, Japan\\\texttt{aebihara@nec.com}}

\begin{document}
%
\maketitle
\begin{abstract}
    Theoretically-inspired sequential density ratio estimation (SDRE) algorithms are proposed for the early classification of time series. Conventional SDRE algorithms can fail to estimate DRs precisely due to the internal overnormalization problem, which prevents the DR-based sequential algorithm, Sequential Probability Ratio Test (SPRT), from reaching its asymptotic Bayes optimality. Two novel SPRT-based algorithms, \texttt{B2Bsqrt-TANDEM} and \texttt{TANDEMformer}, are designed to avoid the overnormalization problem for precise unsupervised regression of SDRs. The two algorithms \textit{statistically significantly} reduce DR estimation errors and classification errors on an artificial sequential Gaussian dataset and real datasets (SiW, UCF101, and HMDB51), respectively. The code is available at: \url{https://github.com/Akinori-F-Ebihara/LLR_saturation_problem}.
\end{abstract}

\section{Introduction}
\label{sec:intro}
\begin{figure}[tb]
\centerline{\includegraphics[width=\columnwidth,keepaspectratio]{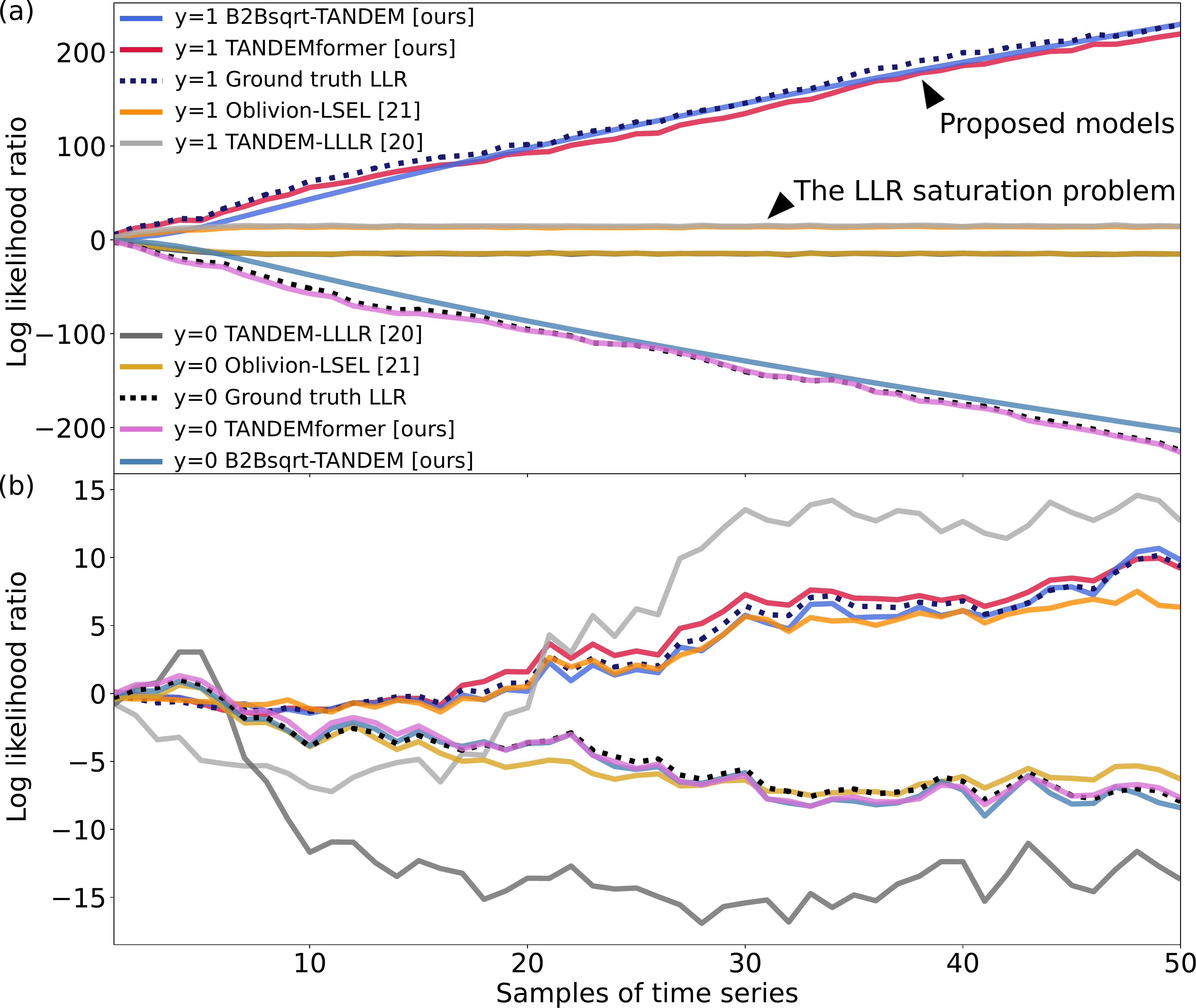}}
\caption{Log-likelihood ratios (LLR) and sequential density ratio estimation (SDRE) results. The LLRs contrast classes $y=1$ and $y=0$: positive and negative LLRs indicate a higher likelihood for class 1 and 0, respectively. Conventional SDRE algorithms show the LLR saturation problem if the LLR's absolute values are large (a) but not if they are small (b).}
\label{fig:killer}
\end{figure}

Early classification of time series \cite{Xing2009ECTSorig, Mori2016reliable_ECDIRE_non-deep_famous, gupta2020approaches_ECTS_recent_review, Hartvigsen2022StopHopEC} is a pivotal algorithm, especially when sampling cost is high, e.g., medical early diagnosis \cite{vats2016early}, autonomous driving \cite{dona2020MSPRT_autonomous_driving}, and action recognition \cite{Weng2020EarlyAR}. Under these applications, the early classifier seeks to optimize both speed and accuracy at the same time. Among the existing early classification methods, the log-likelihood ratio (LLR, i.e., log-density ratio) based algorithm, Sequential Probability Ratio Test (SPRT), is theoretically proven to reach asymptotic Bayes optimality under realistic non-independent and identically distributed (i.i.d.) datasets with more than two classes \cite{wald1947book,TartarBook}. The asymptotic optimality implies that the SPRT algorithm arrives at optimality where the number of samples is minimized under a certain error rate as data is collected. Theoretical study proves that the requirement ensuring the asymptotic optimality is that the absolute values of the LLRs must be increasing \cite{Tartakovsky1998} as more samples are acquired. 
\begin{figure*}[t]
\centerline{\includegraphics[width=\textwidth,keepaspectratio]{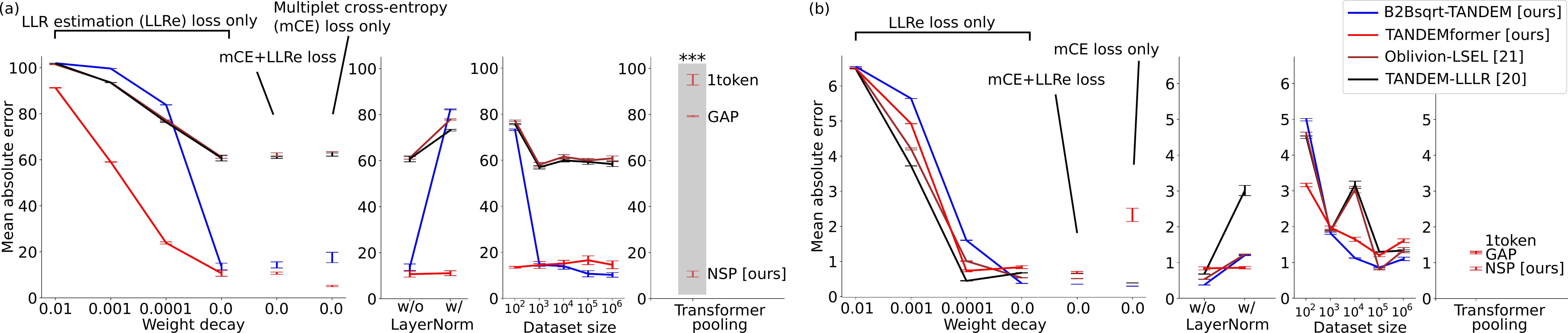}} 
\caption{LLR estimation errors of SDRE algorithms with different weight decay, loss function, normalization, dataset size, and pooling. Each experiment is repeated 10 times to evaluate its standard error of the mean (SEM), plotted as an error bar. Gray shaded areas with asterisks highlights the conditions where the errors are \textit{statistically significantly} reduced (Tukey-Kramer test, $p<0.001$) by the proposed algorithms. Estimation results with LLRs of large (a) and small (b) absolute values are shown.}
\label{fig:MABS}
\end{figure*}

The challenge of applying the SPRT is that the true LLRs of time series are often unknown and must be estimated \cite{sugiyama2008direct, sugiyama2012Density_Ratio_Estimation_in_Machine_Learning, Moustakides2019TrainingNN} with a sequential density ratio estimation (SDRE) algorithm.
We find that the existing SDRE algorithms fail to estimate LLRs when the absolute values of LLRs are large, violating the indispensable condition for the optimality. Specifically, the estimated LLRs 
saturate at their upper limits, accumulating estimation errors as more samples are collected. We define this problem as the \textit{LLR saturation problem} (Fig. \ref{fig:killer}a) and point out that excessive regularization and normalization of internal variables, or \textit{overnormalization}, is the problem source.
The commonly used Long short-term memory (LSTM, \cite{LSTM}) and Transformer \cite{vaswani2017NIPS_Transformer_origianal} can cause the LLR saturation problem because of the overnormalization. Moreover, conventional training tricks such as weight decay \cite{Krogh1991}, Layer Normalization \cite{LayerNorm}, and Global Average Pooling \cite{Lin2014NetworkIN} can also deteriorate the LLR saturation problem. 
Note that the LLR saturation problem essentially differs from the known density chasm problem \cite{TelescopingDRE}: although both problems are phenomenologically similar, the density chasm problem is mitigated in proportion to the dataset size, while the LLR saturation problem is not (Fig. \ref{fig:MABS}a).

Based on the classification-based SDRE algorithm SPRT-TANDEM \cite{SPRT-TANDEM, MSPRT-TANDEM}, we avoid the overnormalization to solve the LLR saturation problem with the two proposed architectures, \texttt{B2Bsqrt-TANDEM} and \texttt{TANDEMformer}. The former has an LSTM backbone equipped with a novel activation function, \texttt{B2Bsqrt}, which is designed to achieve both stable training and precise SDRE. The latter has a streamable Transformer backbone equipped with a novel pooling layer, Normalized Summation Pooling \texttt{(NSP)}, to accumulate tokens that are mixed with self-attention blocks. Note that this is the first paper that applies the Transformer backbone to the early classification problem to the best of our knowledge.

SDRE and early classification experiments are conducted on an artificial sequential Gaussian dataset (i.e., with known ground-truth LLRs) and real datasets (SiW \cite{SiW}, UCF101 \cite{UCF101}, and HMDB51 \cite{HMDB51}), respectively. Tukey-Kramer test 
confirms that our proposed algorithms outperform baselines with \textit{statistically significantly} better performance metrics. This result indicates that our proposed algorithm approaches the theoretically guaranteed optimality, given the assumption that the true LLRs of the real data increase as more samples are collected. 
\section{Method}\label{sec:method}
\begin{figure*}[t]
\centerline{\includegraphics[width=1.0\textwidth, keepaspectratio]{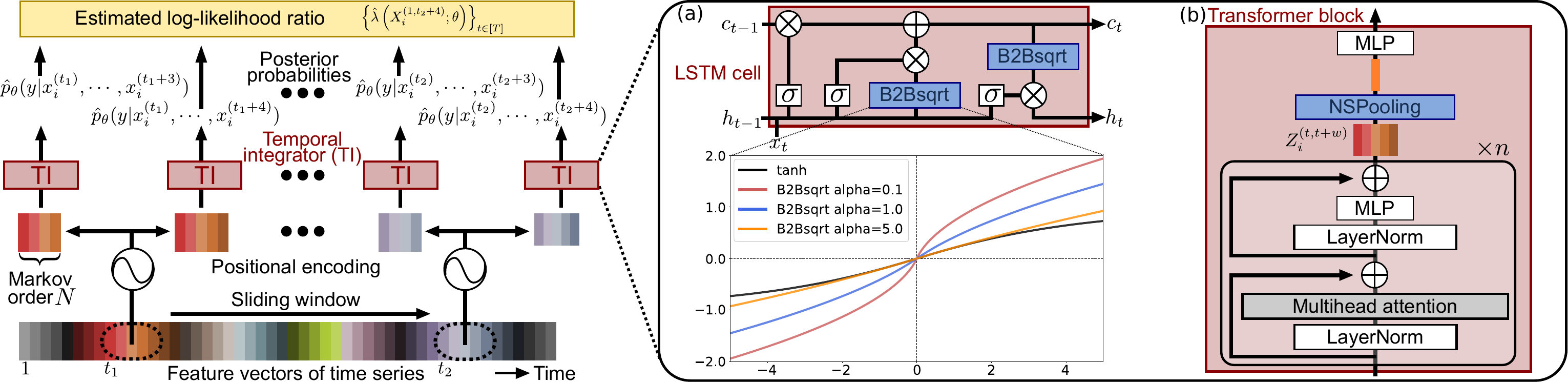}}
\caption{The design of LSTM-based \texttt{B2Bsqrt-TANDEM} (a) and Transformer-based \texttt{TANDEMformer} (b) temporal integrator (TI) for SDRE. A subset of feature vectors of time series is extracted with a sliding window, whose size is defined with SPRT-TANDEM's Markov order $N$ (Thm. \ref{thm:TANDEM_formula}). In the above example, $N=4$. The extracted subset is then processed with the TI to compute posterior probabilities, with which the estimated LLR is computed according to the TANDEM formula (Thm. \ref{thm:TANDEM_formula}).}
\label{fig:Concept}
\end{figure*}

\subsection{Problem setting and notations} Let $X^{(1,T)}_i$ and $y \in [K] := \{1, 2, \dots K\}$ be sequential data $(X^{(1,T)}_i:=\{x_i^{(t)}\}^T_{t=1})$ and a class label, respectively. $i \in [M]$ and $M \in \mathbb{N}$ is the dataset size. $T\in \mathbb{N}$ is the maximum length of the time series, and $x^{(t)} \in \mathbb{R}^{d_{\mathrm{feat}}}$ is a feature vector with length $d_{\mathrm{feat}}$. LLR contrasting class $k$ and $l$ is defined as $\lambda_{kl}(T):=\lambda_{kl}(X^{(1,T)}_i):=\log (p(X^{(1,T)}_i|y=k) / p(X^{(1,T)}_i|y=l))$. Given trainable parameters $\theta \in \mathbb{R}^{d_{\theta}} (d_{\theta} \in \mathbb{N})$, the estimated LLR is denoted as $\hat{\lambda}(X^{(1,T)}_i;\theta)$. The goal of the early classification is to find the correct label $y$ with the smallest possible subset of sequential data $X^{(1, \tau)}, (1 \leq \tau \leq T, \tau \in \mathbb{N})$. 
\subsection{Decision rule} In order to halt data sampling to make an early decision, an early classification algorithm needs to have a stopping rule or a decision rule $\delta$. We select the SPRT, $\delta^*$, as the stopping rule because the SPRT is theoretically proven to be Bayes optimal under i.i.d. data $X^{(1,T)}_i$ with binary classes $y \in [0, 1]$, or asymptotically optimal under non-i.i.d. data with multi-class. 
\begin{definition}[\textbf{SPRT \cite{wald1947book}}] 

Let $d^*$ and $\tau^*$ be the decision rule and stopping time, respectively. Given the decision threshold, $a_{kl} \in \mathbb{R}$, ($k, l \in [K]$) SPRT $\delta^*$ is defined as $\delta^* := (d^*, \tau^*)$, where $d^* := k$ if $\tau^* = \tau_k$ ($k \in [K]$). Here,
    $\tau^* := \underset{k^\prime \in [K]}{\min} \{ \tau_{k^\prime}\}$, and
    $\tau_{k^\prime} := \inf \{ 
        t \geq 1 
        | \underset{l (\neq k^\prime) \in [K]}{\min} \{
            \lambda_{k^\prime l}( X^{(1,t)}_i ) - a_{k^\prime l}
         \geq 0
    \}$.
\end{definition}
Intuitively, the SPRT updates the LLR every time a new sample is acquired until the stopping time when the minimum LLR contrasting a class and others reaches the class threshold.
\subsection{Asymptotic optimality}
\begin{theorem}[\textbf{Asymptotic optimality of the SPRT under a multi-class, non-i.i.d. case \cite{TartarBook}}\label{thm: Asymptotic optimality}]
    Assume that a non-negative increasing function $\psi(t)$ ($\psi(t)\xrightarrow{t\rightarrow\infty} \infty$) and positive finite constants $I_{kl}$ ($k,l \in [K]$, $k \neq l$) exist, such that for some $r > 0$, $\lambda_{kl}(t) / \psi(t) \xrightarrow[t\rightarrow\infty]{P_k\textrm{-}r\textrm{-}quickly} I_{kl}$. Then for all $m \in (0, r]$ and $k \in [K]$, $\underset{\delta}{\inf}\:\mathbb{E}_k[\tau]^m \approx \mathbb{E}_k[\tau^*]^m$ as $\underset{k,l}{\max}\:a_{kl} \rightarrow \infty$.
    
\end{theorem}
The precise definition of $r$-quick convergence and a more detailed discussion can be found, e.g., in \cite{TartarBook}. Intuitively, Thm. \ref{thm: Asymptotic optimality} tells the following: assuming that the LLRs $\lambda_{kl}$ increase as samples are accumulated, the SPRT algorithm reaches the asymptotic optimality where the moments of the stopping time are minimized up to order $r$, given a classification error rate. 

The experiments with real datasets are performed under the assumption that the unknown true LLRs satisfy Thm. \ref{thm: Asymptotic optimality}. This is a natural assumption, given that the features of real scenes continuously bring new information for the classification task.

\subsection{Sequential Density Ratio Estimation}
The following TANDEM formula (Thm. \ref{thm:TANDEM_formula}) is computed using posterior probabilities which are outputs from a temporal integrator network (Fig. \ref{fig:Concept}). The network is optimized with the LLR estimation (LLRe) loss, log-sum exponential loss (LSEL, Def. \ref{def:LSEL}) function, often linearly combined with the multiplet cross-entropy (mCE) loss \cite{SPRT-TANDEM}. In the experiments, the expected value of LSEL is calculated as an empirical mean across timestamps $t$, classwise examples $i$, and classes $k$.
\begin{theorem}[\textbf{TANDEM formula \cite{SPRT-TANDEM}}] \label{thm:TANDEM_formula}
Assuming that $X^{(1,T)}_i$ are $N$th order Markov series, $\lambda_{kl}(t)$can be approximated as:
\begin{align}
\hat{\lambda}_{kl}(X^{(1,t)}_i) &=\log \left(
        \frac{p(x_i^{(1)}, ..., x_i^{(t)}| y=k)}{p(x_i^{(1)}, ..., x_i^{(t)}| y=l)}
    \right)\nonumber \\ 
    = &\sum_{s=N+1}^{t} \log \left( 
        \frac{
            p(y=k| x_i^{(s-N)}, ...,x_i^{(s)})
        }{
            p(y=l| x_i^{(s-N)}, ...,x_i^{(s)})
        }
    \right) \nonumber \\
    - &\sum_{s=N+2}^{t} \log \left(
        \frac{
            p(y=k| x_i^{(s-N)}, ...,x_i^{(s-1)})
        }{
            p(y=l| x_i^{(s-N)}, ...,x_i^{(s-1)})
        }
    \right) 
\end{align}
\end{theorem}

\begin{definition}[\textbf{LSEL \cite{MSPRT-TANDEM}}] \label{def:LSEL}

\begin{align} 
    \hat{L}_{\rm \text{LSEL}} (\bm{\theta}; S) 
    := \mathbb{E} \left[ \log \left( 
        1 + \sum_{ l ( \neq k ) } e^{ - \hat{\lambda}_{k l} ( X_i^{(1,t)}; \bm{\theta} ) }
    \right) \right]. 
\end{align}
\end{definition}

\begin{figure*}[ht] 
\centerline{\includegraphics[width=\textwidth,keepaspectratio]{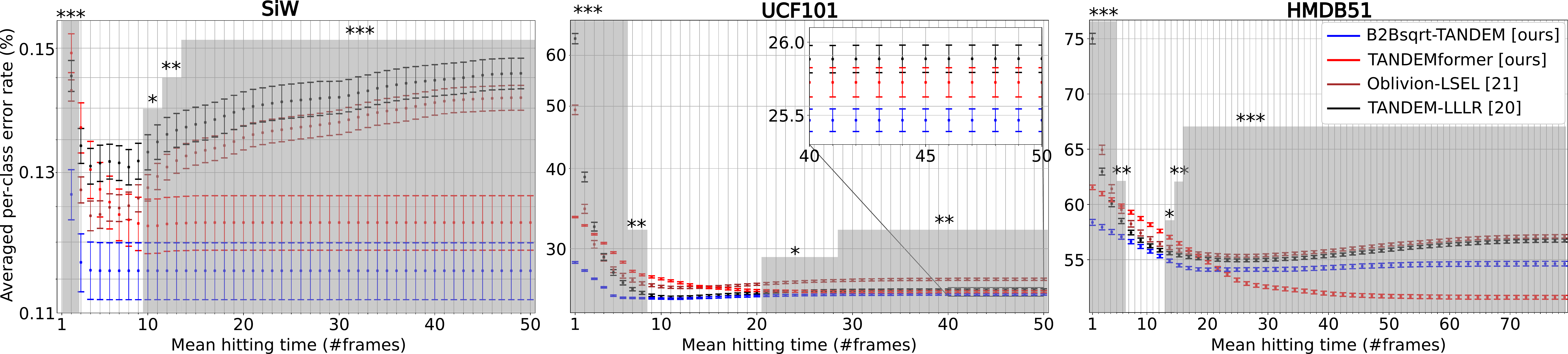}}
\caption{Speed-accuracy tradeoff (SAT) curves to evaluate the early classification performance on the three real databases, SiW, UCF101, and HMDB51. The gray shaded areas show the region where at least one of the proposed models achieves \textit{statistically significantly} better per-class error rate than all the baselines. Maximum $p$-values calculated with the dominant model are shown (Tukey-Kramer test). One, two, and three asterisks show $p<0.05$, $p<0.01$, and $p<0.001$, respectively. Error bars are SEMs.}
\label{fig:SAT}
\end{figure*}
\subsection{\texttt{B2Bsqrt-TANDEM} and \texttt{TANDEMformer}}
To prevent the overnormalization problem, each of the two proposed networks has a simple but effective trick. The \texttt{B2Bsqrt-TANDEM} (Fig. \ref{fig:Concept}a) uses a novel \texttt{B2Bsqrt} function (Def. \ref{def:B2Bsqrt}) in place of the default tanh activation function. 
\begin{definition}\label{def:B2Bsqrt}
\textbf{Back-to-back (B2B) sqrt.} Let $\alpha \geq 0$ as a hyperparameter. The B2Bsqrt function is defined as: 
\begin{align} 
f_{\mathrm{B2Bsqrt}}(x) := \mathrm{sign}(x)(\sqrt{\alpha+|x|}-\sqrt{\alpha})
\end{align} 
\end{definition}
The \texttt{B2Bsqrt} has an unsaturated range $[-\inf, \inf]$ and a finite gradient $1/(2\sqrt{\alpha})$ at the origin to achieve both stable optimization and effective SDRE. We find that $\alpha=1$ is the best value achieving both stability and accuracy, and thus fix the value hereafter. Note that although the \texttt{B2Bsqrt} shares the spirit with the logarithmic activation function \cite{LSTMactivations}, we find that the logarithmic activation can cause unstable training results: thus, we focus on the \texttt{B2Bsqrt} to explore its effect.

\texttt{TANDEMformer} (Fig. \ref{fig:Concept}b) is a streamable transformer with a novel \texttt{NSP} layer. Tokens are first extracted with a sliding window and mixed with self-attention. The \texttt{NSP} then takes a sum of the tokens and divides them with a constant, which is the maximum size of the sliding window. The sliding window size is determined with the order of Markov assumption, $N$ (Thm. \ref{thm:TANDEM_formula}). While GAP and one-token Transformer (e.g., ViT \cite{ViT}) generate one summarized token of similar range irrespective of sample history, \texttt{NSP} can allow evidence accumulation to let LLRs grow without overnormalization (Fig. \ref{fig:MABS}). 
\begin{definition}\label{def:NSP}
\textbf{NSP.} Let $X_i^{(t, t+w)}$ be subtokens sampled with a sliding window of size $w \in [N]$, and let $Z_i^{(t, t+w)}:=\{z_i^{(s)}\}^{t+w}_{s=t}$ be the subtokens mixed with self-attention. Given the Markov order $N$, the \texttt{NSP} layer is defined as:
\begin{align}
NSP(Z_i^{(t, t+w)}) := \sum_{s=t}^{t+w}\frac{z_i^{(s)}}{N+1}.
\end{align}
\end{definition}

\section{Experiments and Results}\label{sec:experiments}
The primary purpose of this experiment is to have a \textit{fair} comparison to outperform the baselines with our algorithms: as long as the comparison is \textit{fair}, we do not try all the available options because it will not affect our conclusion. To achieve this goal, we set a common batch size, feature extractor, feature vector size $d_{\mathrm{feat}}$, Markov order $N$, and LSTM's hidden layer size for all the models. All other hyperparameters are optimized with Optuna \cite{Optuna}. We make our GitHub project page publicly available for reproducibility: the code and all the detailed experimental setups can be found there.
\subsection{Compared models}
Two state-of-the-art SDRE algorithms are used as baselines: SPRT-TANDEM with LLLR \cite{SPRT-TANDEM} and TANDEM with Oblivion and LSEL \cite{MSPRT-TANDEM}. Both adopt the LSTM and TANDEM formula (Thm. \ref{thm:TANDEM_formula}), but the former applies the LLLR loss function \cite{SPRT-TANDEM} for SDRE, while the latter uses LSEL (Def. \ref{def:LSEL}). The latter also uses a simplified LLR estimation formula, Oblivion: $\hat{\lambda}_{kl}(X^{(t-N,t)};\theta)\approx \log{(p(k|X^{(t-N,t)})/p(l|X^{(t-N,t)}))}$ Hereafter the two baselines are abbreviated as TANDEM-LLLR and Oblivion-LSEL, respectively. 
\subsection{Simulated Gaussian datasets with known LLRs}
Let $p_0(x)$ and $p_1(x)$ be the 128-dimensional Gaussian densities with covariance identity and mean $(a, 0, 0, ..., 0)$ and $(0, a, 0, ..., 0)$. $a \in \{1.0, 2.0\}$ is the density offset for simulating LLRs with small and large absolute values. Samples of length $T=50$ are randomly taken from the two separated Gaussians to create training, validation, and test datasets with 80K, 10K, and 10K data, respectively. Additional 920K training data are used only for the large-dataset experiment. SDRE evaluation metric is the mean absolute error between the ground-truth and estimated LLRs. The Markov order is set to the maximum 49. No mCE loss is used for optimization. 

Fig. (\ref{fig:MABS}) shows that all the models increase errors as weight decay increases. With small weight decay parameters, the proposed models show \textit{statistically significantly} smaller estimation errors when $\alpha$=2.0. Layer Normalization also deteriorates the results, except for the \texttt{TANDEMformer}. A larger database size does not mitigate the error, highlighting the challenge of solving the LLR saturation problem. The pooling layer of the Transformer also affects the performance: with the proposed \texttt{NSP} layer, the estimation error is \textit{statistically significantly} reduced compared to the GAP or one-token approach.
\subsection{Real-world datasets for the early classification}
Experiments are repeated 50 times to evaluate the Speed-Accuracy tradeoff (SAT) curve (Fig. \ref{fig:SAT}). Markov order $N$ is fixed to 10. mCE loss is used with LLRe loss weighted with a hyperparameter, LLRe loss ratio. ResNet-152 is trained as a feature extractor for the Spoofing in the Wild (SiW, \cite{SiW}) dataset to generate 512-dim feature vectors. Training, validation, and test datasets contain 46729, 4968, and 43878 videos, and class labels are binary. Pretrained Microsoft Vision Model ResNet50 \cite{Pretrained_ResNet50_microsoftvision} without fine-tuning is used to extract 2048-dim feature vectors from UCF101 \cite{UCF101} and HMDB51 \cite{HMDB51} action recognition datasets. Training/validation/test datasets are 35996/4454/15807, and 1026/106/105 for UCF101 and HMDB51, respectively.  Sample lengths are $T=50$ and $79$. 

In Fig. \ref{fig:SAT}, the algorithm that is as far to the left and as far down in the graph as possible is the better one (i.e., the earlier decision with better accuracy). The gray shaded areas show the regions where at least one of the proposed models shows \textit{statistically significantly} better mean per-class error rates than the baselines. Our models show superior performance at the late phase with larger samples, reminiscent of the asymptotic optimality. Interestingly, some of our algorithms show predominance even at the very early phase, where only a few samples are collected: it probably be a by-product of the precise estimation of the LLRs. The SAT curves of the proposed models mostly show a monotonic decrease as more data are accumulated, indicating that the models successfully accumulate evidence without the LLR saturation problem. Given the increasing LLR assumption, our algorithms are expected to approach the asymptotic optimality (Thm. \ref{thm: Asymptotic optimality}).

\section{Conclusion}

The LLR saturation problem is formulated and solved with highly effective yet simple solutions. They prevent the overnormalization for precise unsupervised regression of the LLRs, providing an essential step toward the asymptotic optimality.  
\clearpage

\bibliographystyle{IEEEbib_AFEmod}
\bibliography{References.bib}

\clearpage
\appendix
\section*{Appendix}
\section{Three-class SDRE}
\begin{figure*}[tb]
\centerline{\includegraphics[width=1.0\textwidth,keepaspectratio]{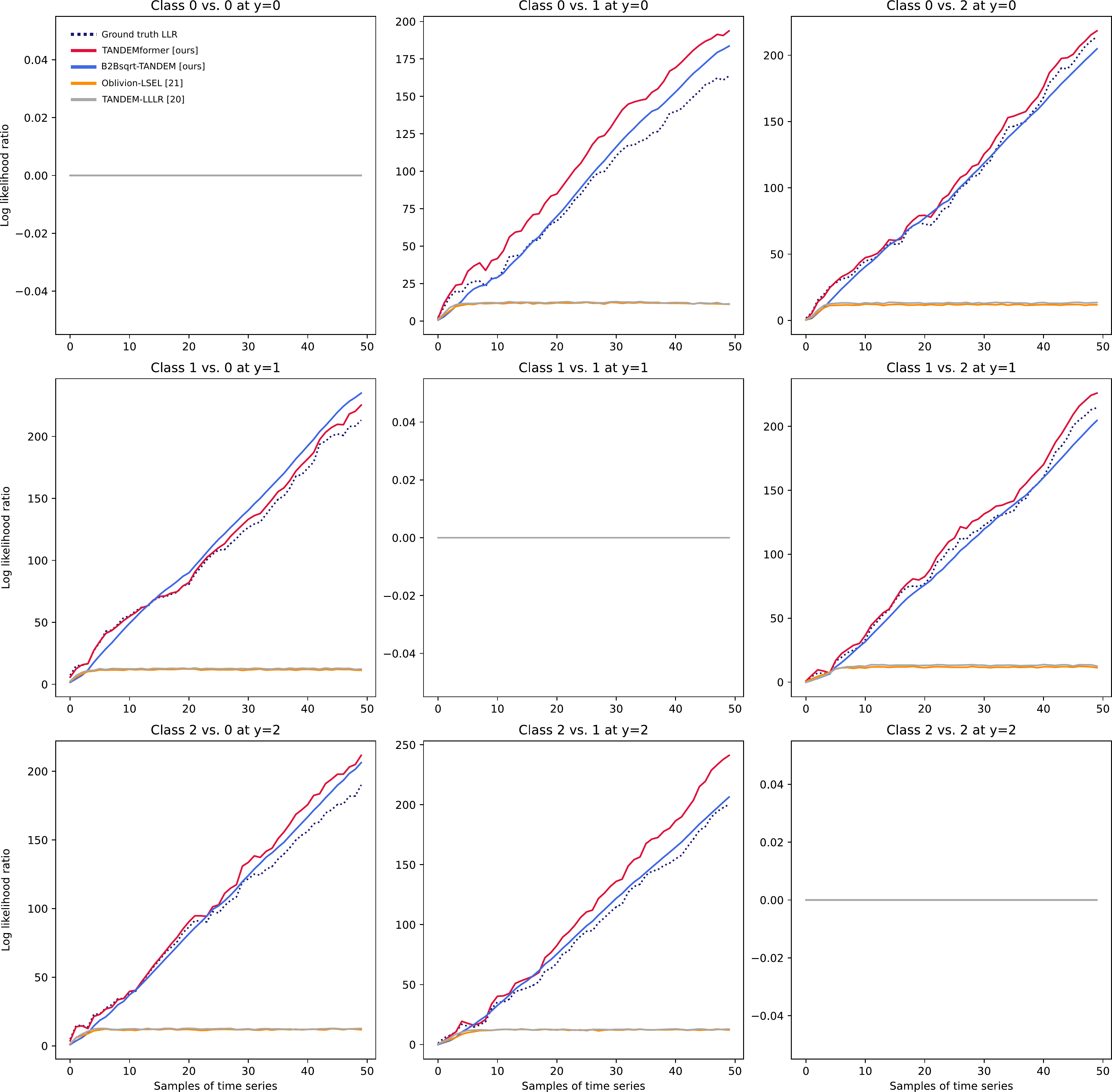}}
\caption{SDRE on a 3-class dataset. "Class $a$ vs. $b$ at $y=a$" indicates that the plotted LLR shows $\log{p(X|y=a) / p(X|y=b)}$, when the ground truth label is $y=a$. The new 3rd class is sampled from a 128-dim Gaussian with mean $(0, 0, 2, ..., 0)$. All other experimental conditions are the same as Fig. 1.}
\label{fig:3classDRE}
\end{figure*}

We could only provide SDRE results with two classes due to the strict page limit of the ICASSP format. As a supplementary experiment, Fig. \ref{fig:3classDRE} shows example results of three-class SDRE. Even with a larger number of classes, the proposed algorithms successfully estimate the LLRs. Thus, our two proposed algorithms, \texttt{B2Bsqrt-TANDEM} and \texttt{TANDEMformer}, effectively estimate the ground-truth density ratio with high precision and can also be used as effective early-classification algorithms on multiclass datasets.

\section{Frequently asked questions}

\noindent\textbf{[How are $T$ and $N$ values selected?]} Both of them can be found empirically with ease. While changing these values may lead to slight variations in the outcome, our algorithms are robust to these hyperparameter changes.

In our previous works \cite{SPRT-TANDEM, MSPRT-TANDEM}, we extensively tested various $T$ from 20 to 100 to find that our algorithm outperformed others. We also deploy commercial products using the proposed algorithms with even shorter $T$ (due to limited computational resources) without problems. Thus, $T$ can be chosen conveniently depending on their application and hardware requirement without sacrificing the performance of our models. However, it is worth noting that the appropriate $T$ should be at least longer than the \textit{specific time scale} described below.

In the prior works, we also extensively discussed choosing $N$ (\cite{SPRT-TANDEM}, Sections 3 and D) and would like to refer interested reviewers to the paper for additional information. In brief, an optimal $N$ can be found either based on the \textit{specific time scale} or hyperparameter tuning. The specific time scale characterizes the data class, e.g., long temporal action such as UCF101 has a long specific time scale, while a spoofing attack such as SiW has a short specific time scale (because one frame can have sufficient information of the attack). Setting $N$ equal to the specific time scale works most of the time. Alternatively, $N$ can be objectively chosen with a hyperparameter tuning algorithm such as Optuna, just as we choose other hyperparameters. Because $N$ is only related to the temporal integrator after feature extraction, the optimization of $N$ is not computationally expensive (3 ($N=0$) to 9 ($N=49$) mins/epoch, \texttt{B2Bsqrt-TANDEM} on SiW with RTX2080Ti).\newline

\noindent \textbf{[Provided comparisons are mainly with the author's prior work.]} The choice of the baseline algorithms is based on two criteria: (1) they are capable of SDRE, and (2) they can be applied to early classification. The state-of-the-art methods that meet these criteria are our prior algorithms. It is worth mentioning that in our two previous papers, we extensively compared our algorithms with existing works by others, including the reinforcement learning algorithm EARLIEST, early classification algorithms LSTM-m and LSTM-s, DRE algorithms LSIF, KLIEP, DSKL, and BARR. Since then, few major SDRE algorithms (if any) for solving early classification have been presented. Thus, our prior algorithms are the most proper ones for the baseline; using them for comparison does not degrade our paper value or affect our conclusion.

\section{Supplementary Discussion}
Our work provides a critical step toward the theoretical asymptotic optimality (Theorem \ref{thm: Asymptotic optimality}) of the SPRT algorithm, which requires precise estimation of LLRs. The asymptotic optimality guarantees reaching the best speed-accuracy tradeoff as more samples are accumulated. As mentioned in Section \ref{sec:method}, the increasing LLR assumption is natural, given that the real scenes continuously bring new information one after another, aiding the classification task. Furthermore, even under violation of the assumption, the SPRT algorithm will often perform at least as equivalent as existing algorithms, if not outperforms. Empirical observations on our in-house data, which contains videos with almost unchanging frames, confirm that our proposed algorithms are on par with the score average of CNN outputs. Thus, our work is expected to benefit the community of DRE and early classification significantly. 

The general challenge of DRE resides in the \textit{large variance} of the estimated density ratio $p_1 / p_0$; i.e., when $p_1$ or $p_0$ is close to zero, the density ratio is almost zero or infinity, respectively. In addition to the variance problem, a seminal paper \cite{TelescopingDRE} found and formulated the \textit{density chasm problem}, where DRE fails counterintuitively even when classification accuracy is perfect. The underlying reason for the density chasm problem remains elusive. Our study uncovered an additional fundamental challenge of DRE: the LLR saturation problem. Our thorough experiments narrowed down the problem to the overnormalization (see Introduction and Fig.2), and we proposed simple but effective solutions to the problem: \texttt{B2Bsqrt-TANDEM} and \texttt{TANDEMformer}. Both are simple and thus easy to implement. 

We expect that our work inspires future works that focus on, e.g., theoretical analysis of the LLR saturation problem and potential relationships between the LLR saturation problem and the density chasm problem.


\section{Author contribution}
A.F.E. conceived the study, conducted the experiments, and wrote the paper. T.M., K.S., and H.I. supervised the study.

\section{Acknowledgments}
The authors thank anonymous reviewers for their careful reading to improve the manuscript. We also thank Hiroshi Fukui for inspiring discussions on Transformer architecture and self-attention.

\end{document}